\documentclass[sigconf,natbib]{acmart}
\usepackage{adjustbox}
\usepackage{balance} 
\usepackage[colorinlistoftodos, textsize=footnotesize]{todonotes}
\AtBeginDocument{%
  }

\setcopyright{acmlicensed}
\copyrightyear{2026}
\acmYear{2026}
\acmDOI{XXXXXXX.XXXXXXX}
\acmConference[XXX '26]{Proceedings of 2026 ACM XXX Conference}{XX, 2026}{XX, XX}
\acmBooktitle{Proceedings of the 2026 ACM XXX Conference, XX, 2026, XX, XX}

\acmISBN{978-1-4503-XXXX-X/2018/06}
\usepackage{placeins}

\begin{document}


\title{Uncovering Political Bias in Large Language Models using Parliamentary Voting Records}


\author{Jieying Chen}
\affiliation{
  \institution{Vrije Universiteit Amsterdam}
  \city{Amsterdam}
  \country{Netherlands}
}
\email{jieying.chenchen@gmail.com}

\author{Karen de Jong}
\affiliation{
  \institution{Vrije Universiteit Amsterdam}
  \city{Amsterdam}
  \country{Netherlands}
}
\email{karenmadejong@gmail.com}

\author{Andreas Poole}
\affiliation{
  \institution{University of Oslo}
  \city{Oslo}
  \country{Norway}
}
\email{andrepoo@math.uio.no}

\author{Jan Burakowski}
\affiliation{
  \institution{University of Amsterdam}
  \city{Amsterdam}
  \country{Netherlands}
}
\email{j.m.burakowski@gmail.com}

\author{Elena Elderson Nosti}
\affiliation{
  \institution{Vrije Universiteit Amsterdam}
  \city{Amsterdam}
  \country{Netherlands}
}
\email{e.i.elderson.nosti@student.vu.nl}

\author{Joep Windt}
\affiliation{
  \institution{Vrije Universiteit Amsterdam}
  \city{Amsterdam}
  \country{Netherlands}
}
\email{j.windt@student.vu.nl}

\author{Chendi Wang}
\affiliation{
  \institution{Vrije Universiteit Amsterdam}
  \city{Amsterdam}
  \country{Netherlands}
}
\email{chendi.wang@vu.nl}



\begin{abstract}
As large language models (LLMs) become deeply embedded in digital platforms and decision-making systems, concerns about their political biases have grown. While substantial work has examined social biases such as gender and race, systematic studies of political bias remain limited—despite their direct societal impact. This paper introduces a general methodology for constructing political-bias benchmarks by aligning model-generated voting predictions with verified parliamentary voting records. 
We instantiate this methodology in three national case studies: PoliBiasNL (2,701 Dutch parliamentary motions and votes from 15 political parties), PoliBiasNO (10,584 motions and votes from 9 Norwegian parties), and PoliBiasES (2,480 motions and votes from 10 Spanish parties). 
Across these benchmarks, we assess ideological tendencies, and political entity bias in LLM behavior. As part of our evaluation framework, we also propose a method to visualize the ideology of LLMs and political parties in a shared two-dimensional CHES (Chapel Hill Expert Survey) space by linking their voting-based positions to the CHES dimensions, enabling direct and interpretable comparisons between models and real-world political actors. 
Our experiments reveal fine-grained ideological distinctions: state-of-the-art LLMs consistently display left-leaning or centrist tendencies, alongside clear negative biases toward right-conservative parties. 
These findings highlight the value of transparent, cross-national evaluation grounded in real parliamentary behavior for understanding and auditing political bias in modern LLMs.
\end{abstract}




\maketitle

\section{Introduction}
The field of natural language processing has recently witnessed rapid advancements with the development of generative models such as GPT and Llama.
As large language models (LLMs) become increasingly integrated across a wide range of global applications, from text rewriting and document summarization to automated customer support and content creation.
Their impact on information dissemination is profound. 
LLMs are also increasingly used as primary sources of information, often replacing traditional search engines~\cite{Kim2023}. 
However, this centralization of information access may restrict diversity, as generative models typically produce only a single synthesized response. This limitation, combined with findings that humans are prone to automation bias~\cite{Simon2020}, raises concerns that model-internal biases may skew public opinion, reinforce stereotypes, and influence decision-making processes unfairly. Therefore, actively detecting and mitigating bias in LLMs is critical to ensuring fairness, trustworthiness, and democratic integrity.

While significant attention has been given to stereotypical biases, such as those related to race and gender, leading to the development of numerous evaluation benchmarks~\cite{parrish-etal-2022-bbq,Mei2023}, political bias remains comparatively underexplored. This is concerning because political bias may exert a stronger influence on users in democratic societies~\cite{Peters2022}. While stereotypical biases are often publicly scrutinized, political biases may be more socially accepted yet equally harmful, especially when embedded in technologies that mediate public discourse.


Existing work on political bias mostly relies on political compass tests and voting advice applications that contain only a few dozen, expert-selected statements tools~\cite{DBLP:conf/acl/RottgerHPHKSH24,DBLP:journals/tacl/CeronFBNP24}. While useful, these instruments are small in scale, subject to selection bias, and fragile under paraphrasing; minor wording can lead to different outcomes, limiting their robustness as benchmarks. They also fall short of the coverage and granularity that is now standard in benchmarks for other types of bias.
In this paper, we address these limitations by constructing a cross-national benchmark for political bias in LLMs grounded in real parliamentary voting records. We align model-generated voting decisions with the documented votes of political parties whose ideological positions are well studied, including through external resources such as the Chapel Hill Expert Survey (CHES)~\cite{2015CHES,CHES}, 
which provides expert assessments of parties on both the economic Left–Right axis and the socio-cultural GAL–TAN axis (Green–Alternative–Libertarian vs. Traditional–Authoritarian–Nationalist).

Concretely, our contributions are fourfold.
(i) We propose a general and robust methodology for constructing political bias benchmarks from parliamentary motions and party votes, and instantiate it for three countries: the Netherlands (2,701 motions, 15 parties), Norway (10,584 motions, 9 parties), and Spain (2,480 motions, 10 parties). 
The datasets are derived entirely from real legislative behaviour and collected through an automated crawling pipeline, enabling longitudinal analyses of ideological drift in LLMs.
(ii) We introduce a language-agnostic evaluation framework that assesses ideological position, and political entity bias in LLM behaviour. As part of this framework, we also propose a method to visualize the ideology of LLMs and political parties in a shared two-dimensional CHES
space by linking their voting-based positions to the CHES dimensions, enabling direct and interpretable comparisons between models and real-world political actors.
(iii) We present a comparative empirical study of widely used LLMs, showing consistent patterns of stronger alignment with left-progressive and centrist parties, along with pronounced negative bias toward right-conservative parties across all three parliaments.
We further verify that these ideological patterns are robust under paraphrased prompt formulations.

\section{Related Work}
  \noindent\textbf{Bias Evaluation Benchmarks.}
A substantial body of work investigates social bias in NLP and LLMs and proposes mitigation strategies across data, modeling, and deployment stages~\cite{Mei2023,gallegos2024biasfairnesslargelanguage}. To quantify such biases, many benchmark datasets have been introduced, including tests for stereotypical associations and question-answering formats that target specific prejudices~\cite{nadeem2020stereosetmeasuringstereotypicalbias,parrish-etal-2022-bbq,mathew2022hatexplainbenchmarkdatasetexplainable}. These benchmarks have been essential for exposing systematic harms, but scores can be sensitive to seemingly minor changes such as negation, paraphrasing, or length, as shown for SocialStigmaQA and related resources~\cite{nagireddy2023socialstigmaqabenchmarkuncoverstigma,selvam-etal-2023-tail}, raising concerns about the robustness and generalization of reported bias metrics.

  \noindent\textbf{Political Bias.}
Political bias in generative language models has mostly been studied using political compass tests and voting advice applications, where models are asked to agree or disagree with a set of expert-curated statements~\cite{feng-etal-2023,Fashoto2024,Motoki2023,DBLP:journals/tacl/CeronFBNP24,DBLP:conf/acl/RottgerHPHKSH24}. These instruments typically contain only 20--65 questions~\cite{DBLP:journals/tacl/CeronFBNP24,DBLP:conf/acl/RottgerHPHKSH24}, limiting scale and granularity compared to social-bias benchmarks that often include hundreds or thousands of examples~\cite{gallegos2024biasfairnesslargelanguage}. They are also vulnerable to paraphrasing effects, with responses changing under minor rephrasing of the same statement~\cite{DBLP:conf/acl/RottgerHPHKSH24}. Beyond the U.S., only a few works examine specific national contexts, for example Dutch or African settings~\cite{Fashoto2024,hartmann2023political}, and these typically rely on small evaluation sets and do not exploit large-scale real-world political decision data.

Political bias has also been shown to affect downstream tasks: partisan training data can lead to divergent performance in hate speech and misinformation detection~\cite{feng-etal-2023}, and opinion summarization models can over-represent left-leaning views~\cite{huang2024biasopinionsummarisationpretraining}. Mitigation strategies, including reinforcement learning from human feedback and related techniques, have been explored but remain only partially effective in aligning models between political perspectives~\cite{Liu2021}.

  \noindent\textbf{Entity Bias.}
 Entity bias refers to systematic differences in model outputs driven by the presence of particular named entities or descriptors rather than by the underlying content of the input~\cite{wang-etal-2023-causal}. Prior work commonly measures such bias using counterfactual prompts in which entities are swapped while the surrounding text is held fixed, and then comparing predictions across such variants~\cite{wang-etal-2022-rely}. Proposed mitigations include masking or perturbing entities during training or inference and adding regularization to encourage invariance to entity substitutions~\cite{Zhu_2022,yan-etal-2022-robustness}.


\section{Cross-National Benchmark Dataset Creation}


Parliamentary votes on motions are a central way for political parties to express their positions. To evaluate political bias in LLMs, we build three benchmarks: \textsc{PoliBiasNL} for the Dutch Second Chamber, containing 2,701 motions, \textsc{PoliBiasNO} for the Norwegian Storting, containing 10,584 motions and \textsc{PoliBiasES} for the Spanish parliament, containing 2480 motions. These benchmarks contain corresponding votes from 15 Dutch, 9 Norwegian and 10 Spanish political parties. By aligning model-generated voting decisions with these recorded party votes, we capture fine-grained ideological signals across political systems.

A political motion is a formal proposal by a parliament member requesting government action or expressing a view on a specific topic.
Manually annotating each motion with an ideology is costly and requires expert knowledge, so we instead exploit the well-documented ideologies of parties via their voting records. This yields a scalable benchmark that spans a broad spectrum of political opinions and can be extended to additional countries and future motions through web scraping.

\subsection{Enhancements}
The current standard practice of using political compass questions to investigate ideological bias in language models~\cite{gallegos2024biasfairnesslargelanguage,DBLP:conf/acl/RottgerHPHKSH24,DBLP:journals/tacl/CeronFBNP24} has several limitations. We address these within the \textsc{PoliBias} benchmarks along three dimensions:

\vspace{-2mm}\paragraph{Diversity and granularity.}
Voting advice tools typically feature only 20--65 statements, whereas bias evaluation datasets often contain hundreds or thousands of examples~\cite{gallegos2024biasfairnesslargelanguage,DBLP:conf/acl/RottgerHPHKSH24,DBLP:journals/tacl/CeronFBNP24}. In contrast, our benchmarks cover thousands of real parliamentary motions from the Netherlands,  Norway and Spain, capturing a broader range of ideological positions than such tools.

\vspace{-2mm}
\paragraph{Selection bias mitigation.}
Political compasses~\cite{DBLP:conf/acl/RottgerHPHKSH24,DBLP:journals/tacl/CeronFBNP24} rely on expert-crafted questions, making selection bias hard to avoid. We reduce this risk by including all motions voted on within a given timeframe, improving representativeness.

\vspace{-2mm}\paragraph{Robustness.}
Minor variations in wording can substantially change model outputs~\cite{selvam-etal-2023-tail}, and political compass benchmarks, with limited variation in their statements, are therefore fragile under paraphrasing. Our benchmarks use naturally authored motions with richer and overlapping semantics, providing a broader range of expressions against which to test models and reducing the influence of any single phrasings on measured political bias.

\subsection{Data Collection}

  \noindent\textbf{\textsc{PoliBiasNL}.} To create the \textsc{PoliBiasNL}  dataset, we developed a custom web scraper 
to extract motions from the Dutch Second Chamber website.\footnote{\href{https://www.tweedekamer.nl/kamerstukken/moties}{https://www.tweedekamer.nl/kamerstukken/moties}} 
The scraper gathered all motions between 2022 and 2024, resulting in a dataset of 2,701 motions.
This period was selected to strike a balance between the political relevance of the data and the breadth needed for comprehensive analysis. 
Additionally, votes from 15 active political parties during this timeframe were collected to establish a baseline for analysis.
We also include essential metadata such as the date, title, motion ID, and both the party names and party members who submitted each motion. 

 %
 \noindent\textbf{\textsc{PoliBiasNO}.} For the Norwegian dataset, we extracted 10,584 political motions submitted to the Storting  from 2018 to 2024.\footnote{\url{https://www.stortinget.no/no/representanter-og-komiteer/partiene/partioversikt/?pid=2017-2021},\url{https://www.stortinget.no/no/representanter-og-komiteer/partiene/partioversikt/?pid=2021-2025}} Similar to the Dutch pipeline, we collected voting records from 9 major Norwegian political parties during the same period. Metadata such as submission date, motion ID, and submitting parties were also collected. The Norwegian motions follow a comparable structure, enabling alignment with our analysis framework.

%
\noindent\textbf{\textsc{PoliBiasES}.}
For the Spanish dataset, we collected official voting records from 2016 to 2025 using a custom web scraper applied to the parliamentary records of 
the Spanish Congress of Deputies (Congreso de los Diputados).\footnote{\url{https://www.congreso.es/es/opendata/votaciones}}
To enable party-level analysis, we aggregated our initial dataset of over 270,000 individual votes into collective positions based on the majority vote within each party. After reviewing duplicate identifiers, we retained entries with distinct dates or records, yielding a final dataset of 2,480 initiatives.

All datasets can be periodically updated to accurately reflect changes in the political landscape by rerunning the scraping code.
\subsection{Data Processing}

A typical political motion includes a title, an introduction or preamble, several recitals outlining considerations, and operative clauses proposing actions. To avoid framing effects, we include only the operative clauses in the benchmark, since the other sections often contain persuasive language that could influence model responses.

Party votes are encoded numerically, with \textbf{1} representing votes in favour and \textbf{–1} representing votes against. Specifically, in the Spanish dataset, \textsc{PoliBiasES}, we additionally include \textbf{0} for abstentions, as abstaining is permitted in the Spanish Congress. 
Over the period covered by our motion dataset, \textit{GL} and \textit{PvdA} merged; when both parties cast the same vote, we retain that value, and when they differed, we assign the outcome 0. The \textit{NSC} party was established partway through this period; therefore, we retrospectively assign earlier votes cast by its leader, Pieter Omtzigt, to \textit{NSC}.

\section{Evaluation}
To more effectively detect political bias within our benchmark, we have designed various experiments aimed at analyzing ideological biases, detecting political entity biases towards specific parties. 
In these experiments, we utilized various prompts paired with specific motions, employing a zero-shot approach to prompt an LLM. 
This paper includes only the English translations of the Dutch, Norwegian and Spanish prompts. 
We evaluate a representative selection of widely used generative LLMs: Mistral-7B~\cite{2310.06825}, Falcon3-7B~\cite{falcon}, Gemma2-9B~\cite{gemma}, Deepseek-7B~\cite{deepseek},
GPT-3.5 Turbo~\cite{openai2023gpt35}, 
 GPT-4o mini~\cite{gpt4}, 
 Llama2-7B~\cite{2023arXiv230709288T} and Llama3-8B~\cite{dubey2024Llama3herdmodels}.
In addition, we evaluate language-specific LLMs trained on particular linguistic domains, including NorskGPT~\footnote{\url{https://www.norskgpt.com/}} for \textsc{PoliBiasNO} and 
Aguila-7B~\footnote{\url{https://huggingface.co/projecte-aina/aguila-7b}} for \textsc{PoliBiasES}.

  %
  \noindent\textbf{Computing Infrastructure.}
For experiments involving open-source LLMs, we used A4000 GPUs via the Hugging Face interface, while GPT models were accessed through the OpenAI API. 
To ensure deterministic outputs across all runs, we set the temperature parameter to 0 for every model evaluated. 

\subsection{Ideological Bias}
To better analyse the ideological stances of LLMs, we evaluate how well these models align with political parties based on their voting choices on motions. 
By comparing the voting results of the LLMs with those of the political parties across a large dataset of motions, we can gauge the extent of their alignment. 
Given that these parties have well-defined political stances, such an examination helps us ascertain the ideological positions of the LLMs more accurately.


Therefore, this experiment aims to identify the ideological orientation of LLMs by presenting the political motions from our benchmark dataset. 
It assesses the model's inherent ideological biases without supplementary context. 
To accomplish this, we ask the LLMs to either support (\textit{`for'}) or oppose (\textit{`against'}) each motion based on a standardized prompt outlined in its  prompt:

 \begin{quote}
    \textbf{System prompt: }\textit{Vote for or against the following motion. only respond with `for’ or `against’.}
\end{quote}

\begin{quote}
    \textbf{User prompt: }\{\textit{motion}\}
\end{quote} 
This version of a prompt is also extended with (\textit{`abstain'}) option for testing against the Spanish benchmark to accurately reflect the real voting environment.








\vspace{1mm}
\noindent\textbf{Projecting LLMs into a Shared CHES Ideological Space.}
Political parties in each country have well-established ideological
positions in the two-dimensional CHES framework~\cite{2015CHES,CHES}, consisting of an
economic left--right axis and a socio-cultural GAL--TAN axis. The CHES
scores for each party are obtained through expert surveys conducted by
political scientists, and are widely used as reference points in
comparative politics. The left--right dimension captures parties'
positions on economic policy, redistribution, and the role of the state
in the economy, whereas the GAL--TAN dimension (`Green--Alternative--
Libertarian' vs.\ `Traditional--Authoritarian--Nationalist') captures
their stances on socio-cultural issues such as immigration, civil
liberties, and cultural values. Our goal is to place LLMs and political parties in the same
space to enable direct comparison between model outputs and real
political actors.

To achieve this, we leverage the fact that both parties and LLMs cast
votes on the same set of parliamentary motions. Parties’ voting patterns
form a matrix $X_{\text{party}}$, and their expert-rated CHES scores
provide the corresponding ideological coordinates $Y_{\text{party}}$.
Recovering CHES positions from voting behaviour can be formulated as a
supervised mapping problem: learning a function $f : X \rightarrow Y$
such that $f(X_{\text{party}}) \approx Y_{\text{party}}$.
We use Partial Least Squares (PLS) regression~\cite{PLS} to estimate this
mapping. 
Unlike Principal Component Analysis (PCA)~\cite{PCA}---which identifies components that explain maximum
variance in voting patterns alone---PLS identifies components that
maximise the covariance between votes and CHES scores, producing
representations that are directly oriented toward ideological structure.
PLS is fit exclusively on the party data, and leave-one-out validation
shows that more than 81--97\% of the variance in the CHES left--right and
GAL--TAN dimensions can be recovered from voting patterns alone. 
This indicates that parliamentary votes contain sufficient information to
approximate parties’ positions in the CHES space.~\footnote{ We also trained a ridge regression model~\cite{ridge} as an alternative
supervised mapping. Ridge achieved similar predictive accuracy and yielded very similar LLM placements.}

Once the model is trained, we compute PLS component scores for each LLM and the new parties without CHES scores (i.e. \textit{JA21} in the Netherlands) based on its voting vector $X_{\text{LLM}}$ These scores are then passed
through the fitted regression functions to obtain predicted CHES
coordinates $(\widehat{\text{LR}}, \widehat{\text{GAL--TAN}})$ for each
model. 
Because the mapping is learned solely from party behaviour and
applied unchanged to LLMs, the resulting coordinates are directly
comparable to the party positions.

Finally, we visualise parties and LLMs in a shared two-dimensional CHES space, providing an interpretable, standardised view of ideological alignment grounded in real voting behaviour.

\begin{figure*}[t]
  \centering

  {\small\textbf{(a) \textsc{PoliBiasNL}}}\\[-1.2em]

  \begin{minipage}[t]{0.30\textwidth} 
    \centering
    \vspace{10pt} 
    \includegraphics[width=0.94\linewidth]{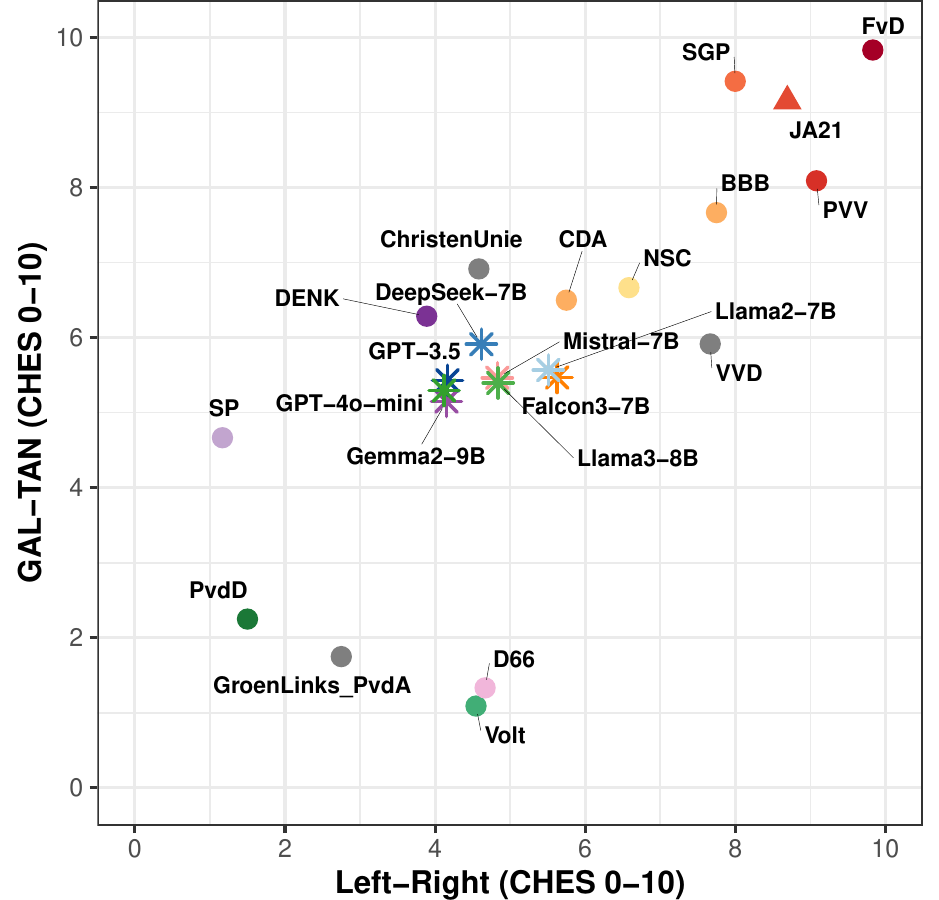}
  \end{minipage}
  \hfill
  \begin{minipage}[t]{0.69\textwidth} 
    \centering
    \vspace{1pt} 
    \includegraphics[width=0.94\linewidth]{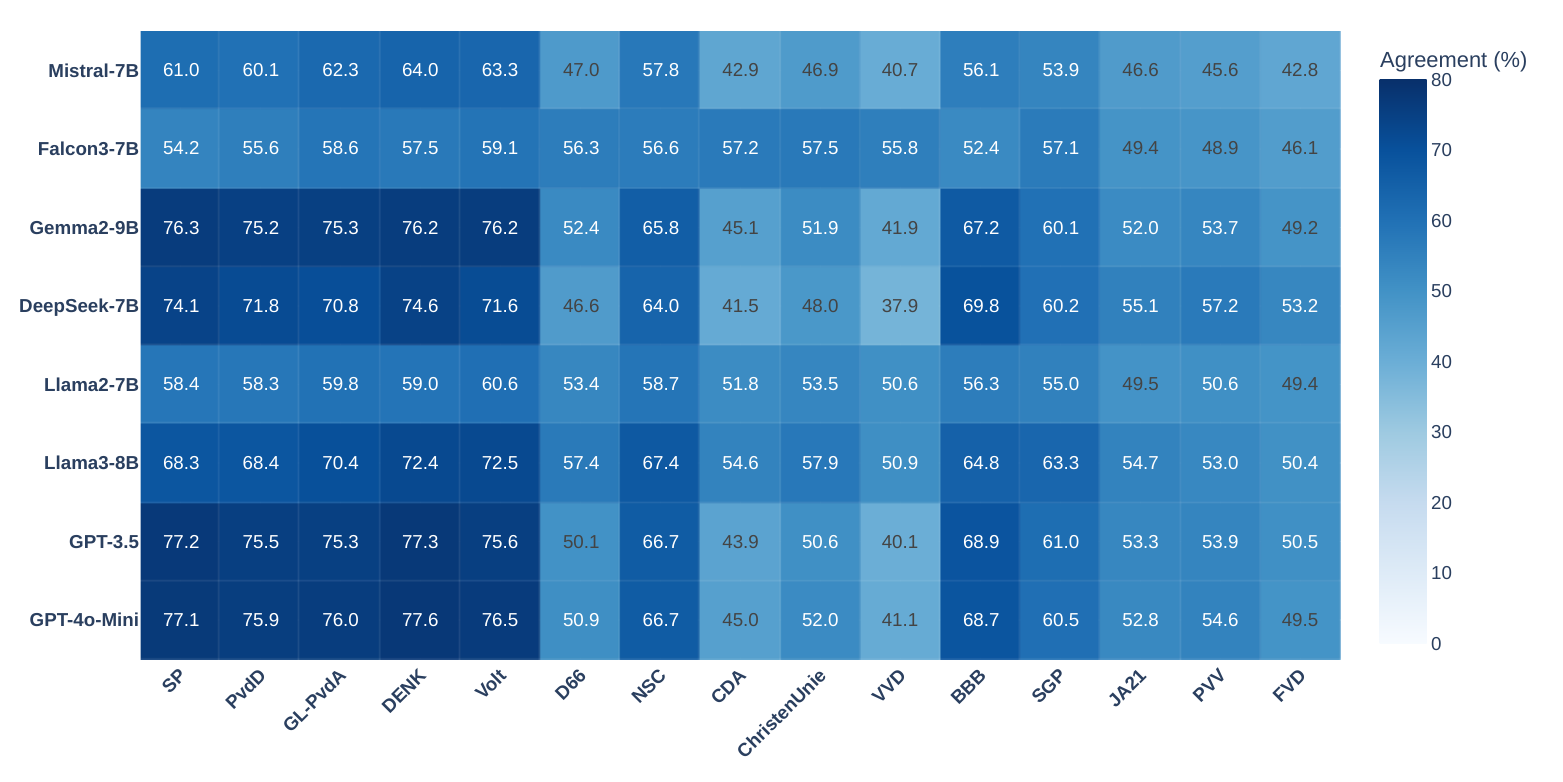}
  \end{minipage}%
\vspace{0.01em}
    {\small\textbf{(b) \textsc{PoliBiasNO}}}\\[-1.2em]
  
  \begin{minipage}[t]{0.30\textwidth} 
    \centering
    \vspace{10pt} 
    \includegraphics[width=0.98\linewidth]{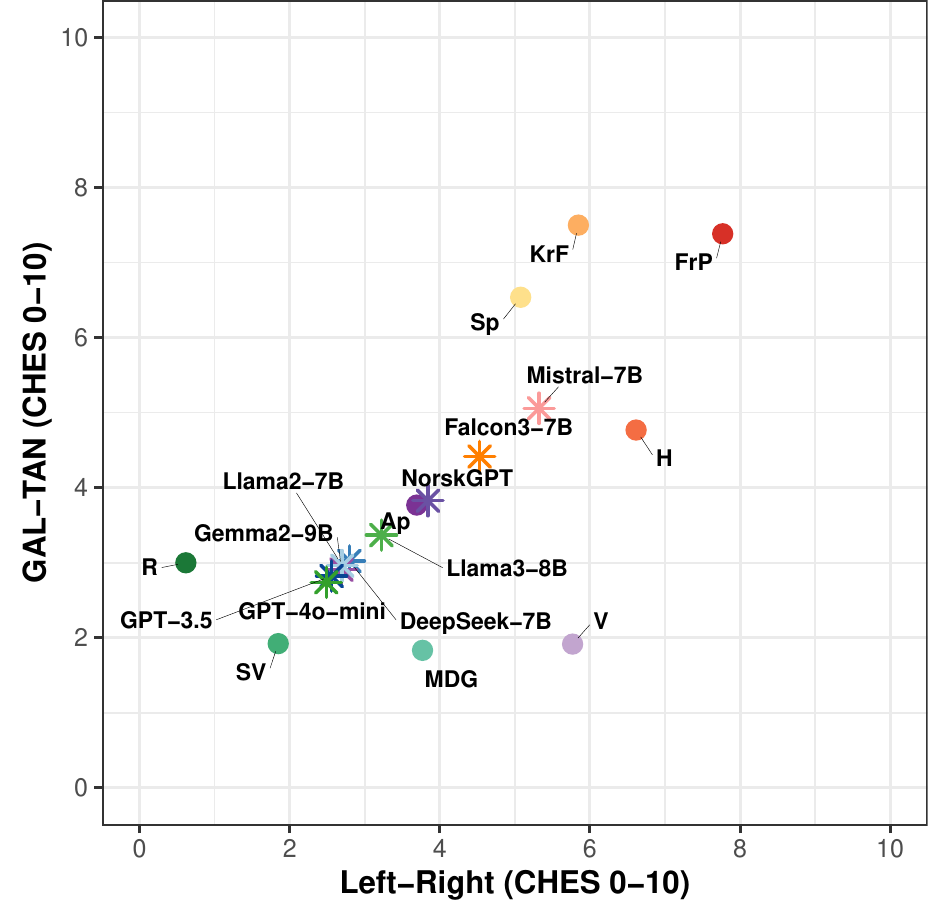}
  \end{minipage}
  \hfill
  \begin{minipage}[t]{0.69\textwidth} 
    \centering
    \vspace{1pt} 
    \includegraphics[width=0.94\linewidth]{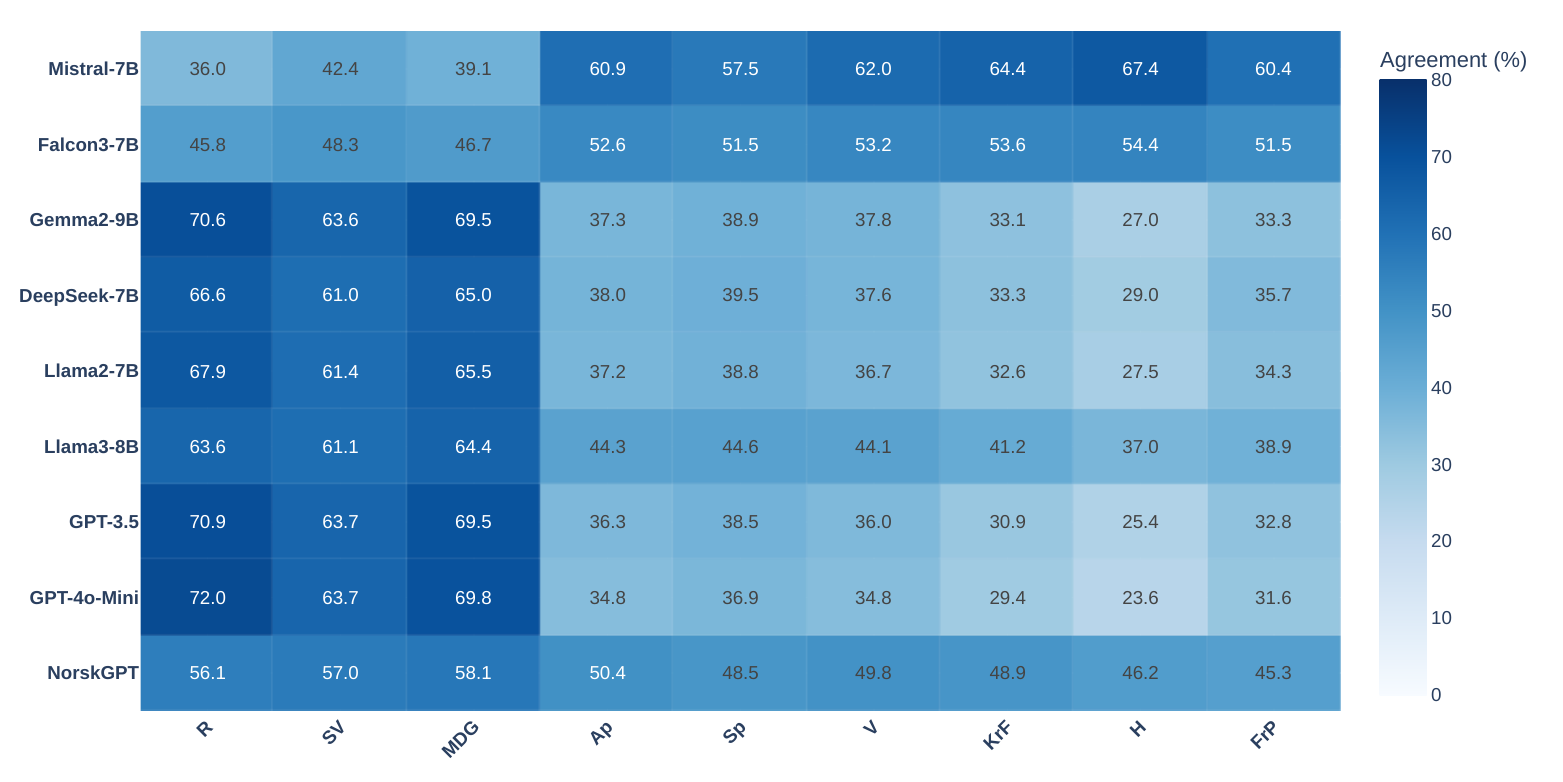}
  \end{minipage}%
  \vspace{0.01em}
{\small\textbf{(c) \textsc{PoliBiasES}}}\\[-1.2em]
    
  \begin{minipage}[t]{0.30\textwidth} 
    \centering
    \vspace{10pt} 
    \includegraphics[width=0.98\linewidth]{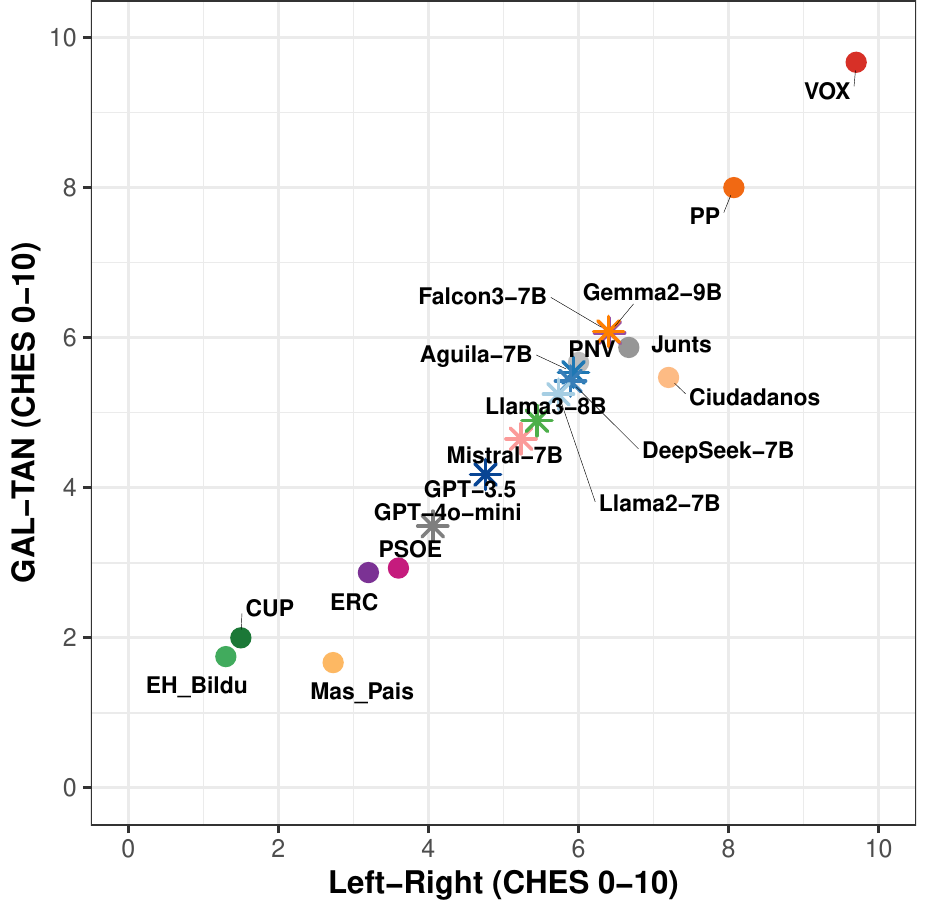}
  \end{minipage}
  \hfill
  \begin{minipage}[t]{0.69\textwidth} 
    \centering
    \vspace{1pt} 
    \includegraphics[width=0.91\linewidth]{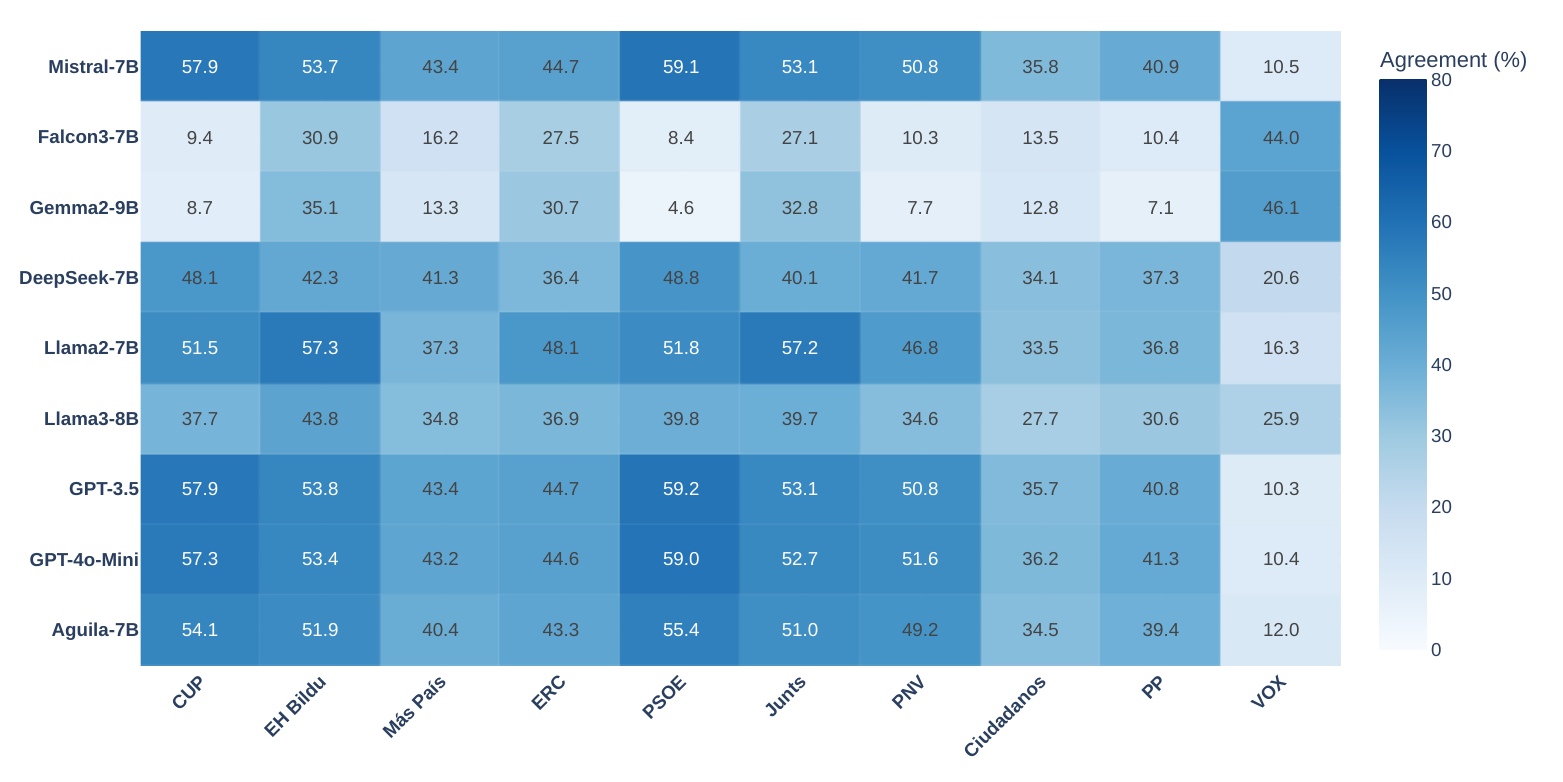}
  \end{minipage}%

    \caption{(Left) Ideological placement of political parties based on the CHES scores in political science, where the Left–Right axis captures economic ideology and the GAL–TAN axis represents socio-cultural values from Green/Alternative/Liberal to Traditional/Authoritarian/Nationalist. (Right) Voting agreement between LLMs and political parties across the three datasets. The parties on the x-axis are ordered from left-progressive to right-conservative ideologies.}


  \label{fig:ideology}
\end{figure*}

\vspace{1mm}
\noindent\textbf{Results: Ideological Positioning of LLMs in the CHES Space.}
The left-hand panels of Fig.~\ref{fig:ideology} show the projected ideological positions of political parties and LLMs in the two-dimensional CHES space for \textsc{PoliBiasNL}, \textsc{PoliBiasNO}, and \textsc{PoliBiasES}. Across all countries, LLMs cluster tightly in the centre–left and moderately GAL-oriented region, exhibiting a consistent ideological footprint.

For the Netherlands (top-left panel), LLMs occupy Left--Right scores of roughly 4--6, i.e., in the same economic band as the progressive and centre-left bloc including \textit{D66} (social-liberal), \textit{GroenLinks--PvdA} (social-democratic/green-left), and \textit{Volt} (pro-EU liberal). 
Along the economic dimension, current LLMs therefore map to the centre-left part of the party system.
However, this alignment does \emph{not} carry over to the socio-cultural GAL--TAN axis. 
Whereas the progressive parties \textit{GL-PvdA}, \textit{PvdD}, \textit{Volt}, and \textit{D66} are located in the strongly GAL-oriented region (scores around 1--2), the LLMs cluster at much higher GAL--TAN values (around 5--6), closer to more moderate or mildly traditional parties such as \textit{DENK}, \textit{ChristenUnie}, and \textit{CDA}. 
For the Norwegian case (middle-left panel), LLMs shift slightly further toward
the left–GAL region compared to their positions in the Dutch CHES space. As a
result, they lie much closer to the core of the Norwegian progressive bloc,
including \textit{Ap} (centre-left social-democratic), \textit{SV}
(left–socialist), \textit{R} (far-left), and \textit{MDG} (green–progressive).
For the Spanish case (bottom-left panel), LLMs exhibit a noticeably different
geometry compared to the Dutch and Norwegian settings. The models form a
strikingly linear cluster running diagonally across the CHES space, with positions
slightly further to the right on the Left--Right axis than in the Dutch and
Norwegian cases, while maintaining comparable values on the GAL--TAN dimension.
As a result, the LLM cluster lies between the moderate left and centrist bloc,
showing closest proximity to \textit{PSOE} (centre-left), \textit{ERC} (left-wing
Catalan nationalist), and \textit{Junts} (centrist Catalan nationalist), while
remaining clearly separated from the mainstream conservative \textit{PP} and the
far-right \textit{VOX}, which occupy the right–TAN extreme. 

Overall, the CHES projections reveal a notable cross-national regularity: LLMs tend to adopt centre-left economic positions and liberal--progressive socio-cultural values, while maintaining clear distance from right-conservative and far-right actors. These projections provide a transparent and interpretable view of model ideology grounded in real parliamentary behaviour.

\vspace{1mm}
\noindent\textbf{Results: Voting Agreement with Political Parties.}
To further validate these patterns using a complementary metric, we next examine direct voting-agreement between models and political parties.
The right-hand panels of Fig.~\ref{fig:ideology} complement the CHES projections by reporting direct voting agreement between LLMs and political parties. Because parties are ordered from left-progressive to right-conservative, the heatmaps provide a structured view of ideological alignment. Across all three countries, the agreement patterns strongly parallel the CHES-based results: LLMs show substantially higher agreement with left-wing, green, and social-democratic parties, and systematically lower agreement with right-conservative and far-right parties.
In the Netherlands (panel~(a)), LLMs reach high agreement with left-progressive parties such as \textit{SP}, \textit{PvdD}, \textit{GL-PvdA}, and \textit{DENK}, but the lowest with the far-right \textit{PVV} and \textit{FvD}. The Norwegian results (panel~(b)) show the same ordering: highest agreement with \textit{R}, \textit{SV}, and \textit{MDG}, moderate alignment with \textit{Ap}, and minimal agreement with \textit{H} and the right-populist \textit{FrP}. In Spain (panel~(c)), models again align most with left-wing parties, and show very low agreement with \textit{PP} and especially \textit{VOX}.

Overall, the heatmaps reinforce the CHES-based interpretation: LLMs systematically resemble the voting behaviour of left-progressive and centre-left parties and diverge sharply from right-conservative blocs. Language-specific models such as NorskGPT and Aguila-7B follow the same pattern, indicating that these ideological tendencies are not tied to a single dataset, language, or model family. The consistency across three parliaments and nine LLMs highlights a stable, cross-national regularity in current-generation LLM behaviour.



\noindent In combination, the CHES projections and agreement heatmaps paint a coherent picture: current-generation LLMs consistently adopt centre-left, liberal-progressive ideological positions across three countries, three languages, and multiple political systems.

 \noindent\textbf{Model Certainty as an Evaluation Metric.}
Existing bias metrics such as those proposed in~\cite{nagireddy2023socialstigmaqabenchmarkuncoverstigma}, capture only the binary response (for or against). 
We aim to address the limitations identified by~\cite{röttger2024political}, which criticize fixed-choice formats for not revealing the strength of model preferences. Inspired by bias metrics used for masked language models~\cite{parra-2024-unmasked,nadeem2020stereosetmeasuringstereotypicalbias}, our metric evaluates the probabilities assigned to each token within the model's responses.

As required by the prompt, the LLM can only respond with \textit{`for'} or \textit{`against'}.
Consequently, we focus solely on the probabilities of the tokens \textit{`for'} and \textit{against'}, as confirmed by our evaluation results where the LLMs consistently produced only these two responses.
For various LLM series, we calculate the probability of these generated tokens differently. 
For the Llama models, we compute the probabilities of the tokens \textit{`for'} and \textit{`against'} by applying the softmax function to the model's logit scores for these tokens. 
In contrast, for the GPT models, this preprocessing step is unnecessary, as the log probabilities for each token can be directly retrieved from the API. These log probabilities are then exponentiated to derive the actual probabilities for those two tokens.

We normalise probabilities of the generated tokens to assess the model's certainty between two choices using the following formula: 
\[
P_{\text{norm}} = \frac{\max(P_{+}, P_{-})}{P_{+} + P_{-}},
\]
where \(P_{+}\) represents the probability of the token \textit{`for'} and \(P_{-}\) denotes the probability of the token \textit{`against'}. 
The normalised probability \(P_{\text{norm}}\) quantifies the model's confidence in choosing between two options. 
It ranges from 0.5 (low certainty) to 1 (high certainty).

\begin{figure}[t!]
    \centering

    {\small\textbf{(a) \textsc{PoliBiasNL}}}\\
    \includegraphics[width=0.95\linewidth]{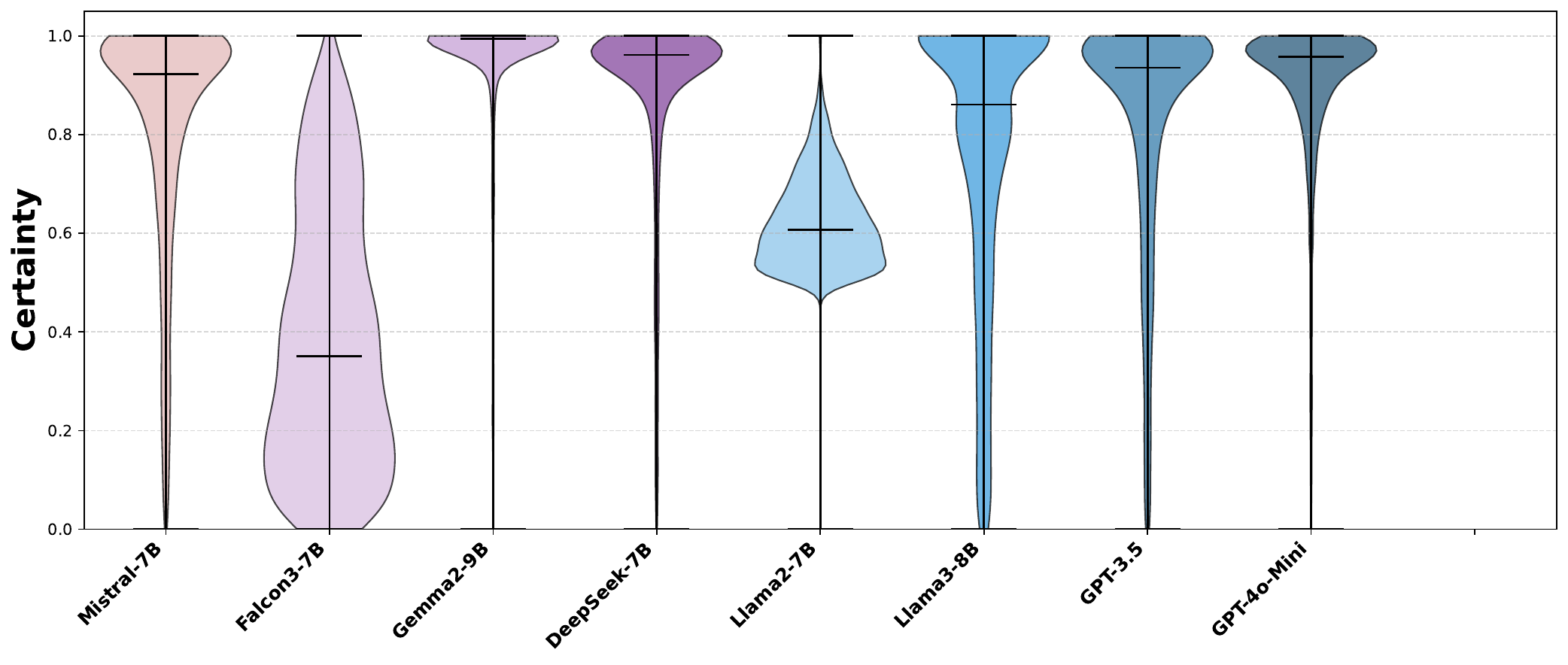}

    {\small\textbf{(b) \textsc{PoliBiasNO}}}\\
    \includegraphics[width=0.95\linewidth]{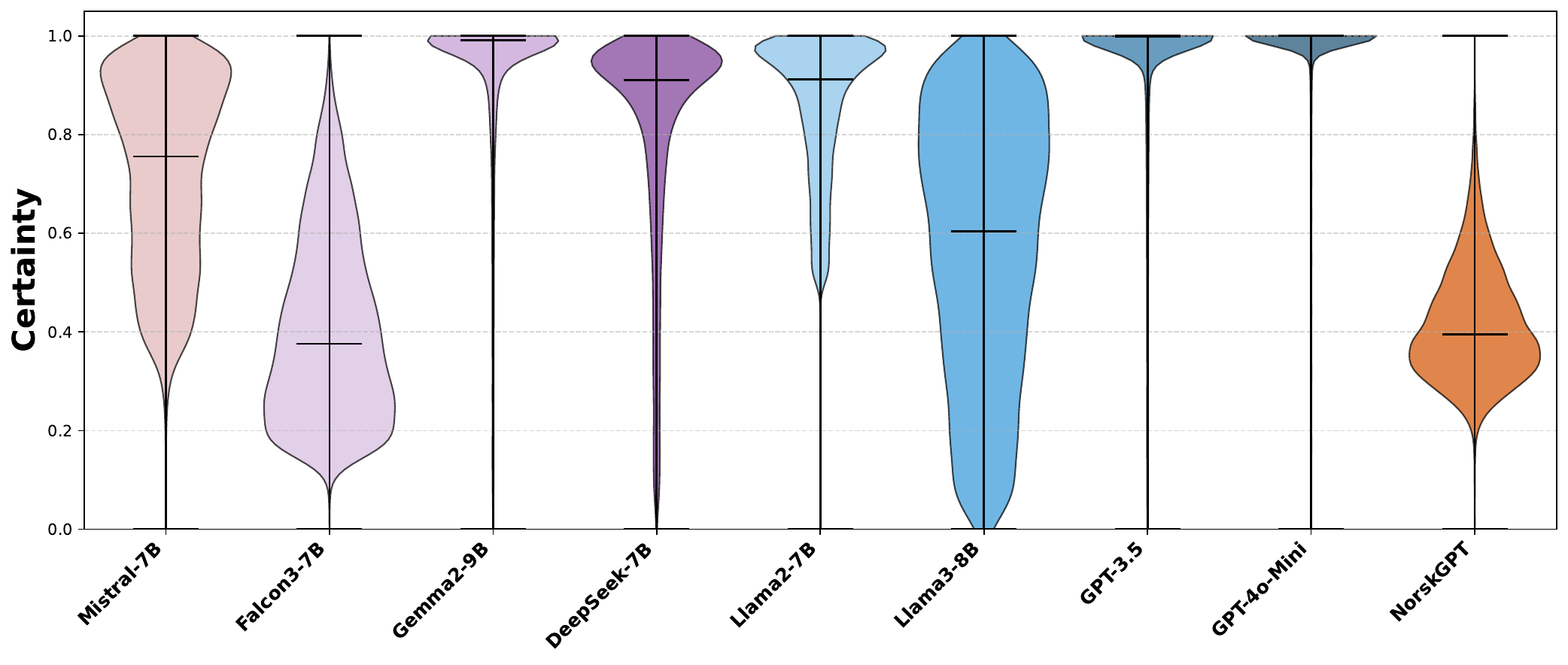}

    {\small\textbf{(c) \textsc{PoliBiasES}}}\\
    \includegraphics[width=0.95\linewidth]{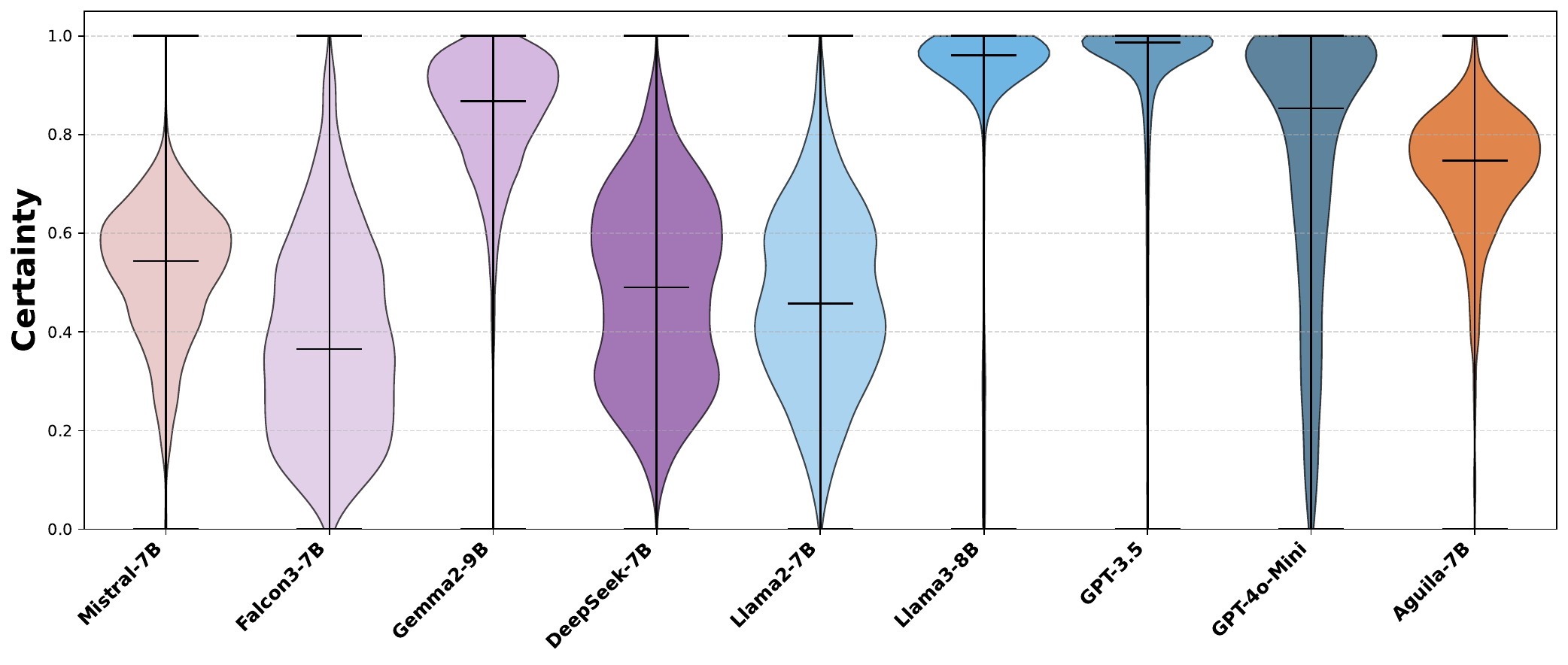}

    \caption{Violin plots showing the distribution of normalised probabilities for the tokens
    \textit{`for'} and \textit{`against'} in response to ideology prompts across models. }
    \label{fig:violin_plot}
\end{figure}

  \noindent\textbf{Results: Ideology Bias – Model Certainty.}
The violin plots in Fig.~\ref{fig:violin_plot} illustrate the distribution of normalised probabilities for the tokens \textit{`for'} and \textit{`against'} in each model’s responses to ideological prompts, reflecting their certainty levels. 
Across the Dutch, Norwegian, and Spanish datasets, three broad patterns emerge.

First, GPT models consistently display the highest certainty, with extremely peaked distributions near~1.0. Both GPT-3.5 and GPT-4o-mini rarely produce low-confidence predictions, suggesting that these models adopt strong and stable ideological commitments. This high certainty aligns with their concentrated placement in the CHES projections, where they exhibit clear centre-left and GAL-oriented ideological positions.

Second, the Llama family shows more variable certainty. Llama3-8B exhibits moderately high confidence across all three parliaments, while Llama2-7B produces substantially flatter distributions, especially in the Dutch and Spanish datasets, indicating more uncertainty and less stable voting behaviour. This variability is consistent with their more spread-out positions in the CHES maps.

Third, other small open models—such as Falcon3-7B, DeepSeek-7B, and Mistral-7B—generally exhibit broader distributions with lower median certainty, highlighting greater indecision or sensitivity to prompt phrasing. 
Language-specific models follow this pattern as well: NorskGPT and Aguila-7B show some improvement on their respective national datasets, but they do not reach the confidence levels of GPT models.

Taken together, these results show a clear link between ideological coherence and model certainty. LLMs that cluster tightly in the CHES ideological space (notably GPT models) also make consistently high-confidence predictions, while LLMs with more diffuse ideological positions exhibit lower and more variable certainty. Certainty thus offers a complementary signal for understanding the stability and reliability of ideological behaviour in LLMs.
We also report the proportion of invalid LLMs outputs in the Appendix~\ref{appendix:invalid}.

\subsection{Entity Bias}

\begin{figure*}[t]
    \centering

    {\small\textbf{(a) \textsc{PoliBiasNL}}}\\[-0.1em]
    \begin{minipage}{0.48\textwidth}
        \centering
        
        \includegraphics[width=\textwidth]{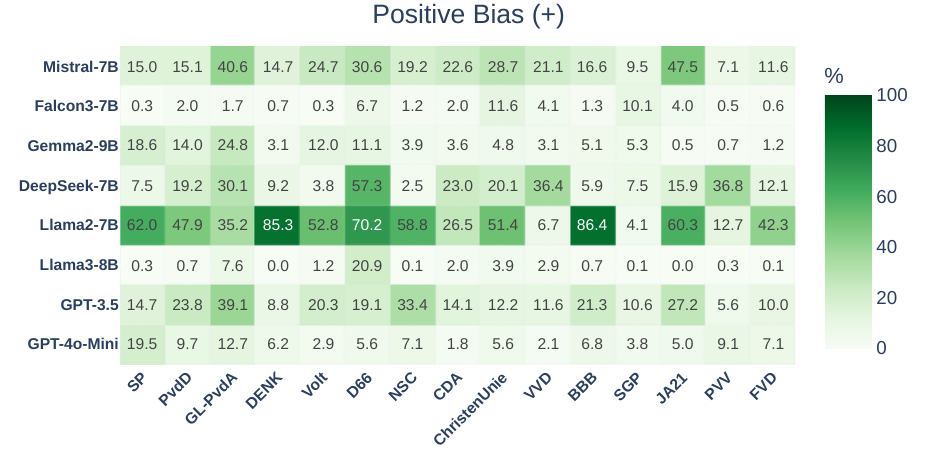}
    \end{minipage}
    \hfill
    \begin{minipage}{0.48\textwidth}
        \centering
        \includegraphics[width=\textwidth]{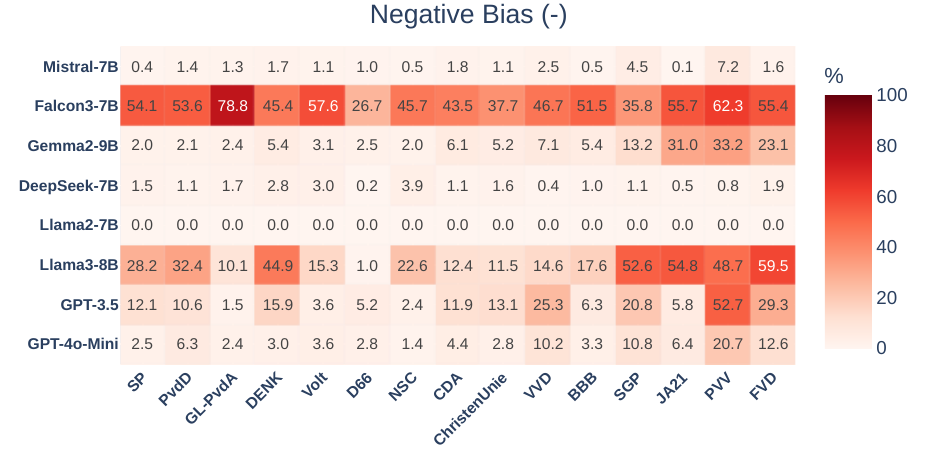}
    \end{minipage}

  {\small\textbf{(a) \textsc{PoliBiasNO}}}\\[-0.1em]
    \begin{minipage}{0.48\textwidth}
        \centering
        
        \includegraphics[width=\textwidth]{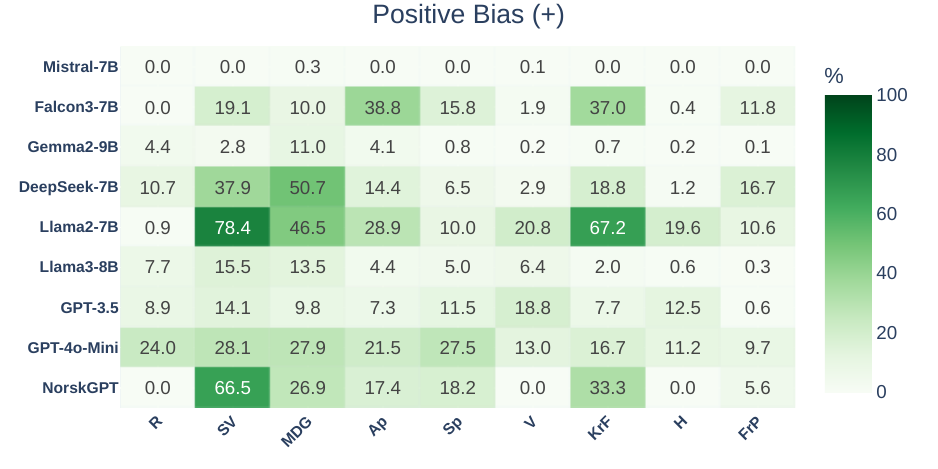}
    \end{minipage}
    \begin{minipage}{0.48\textwidth}
        \centering
        \includegraphics[width=\textwidth]{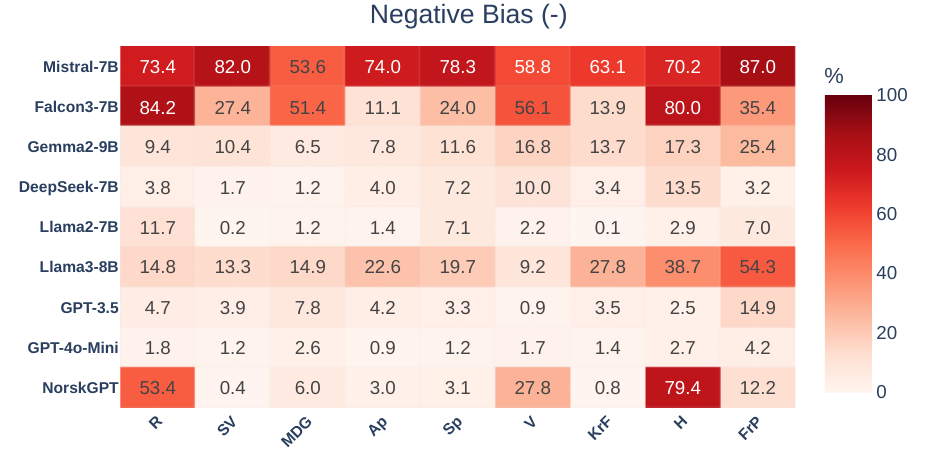}
    \end{minipage}

      {\small\textbf{(a) \textsc{PoliBiasES}}}\\[-0.1em]
    \begin{minipage}{0.48\textwidth}
        \centering
        
        \includegraphics[width=\textwidth]{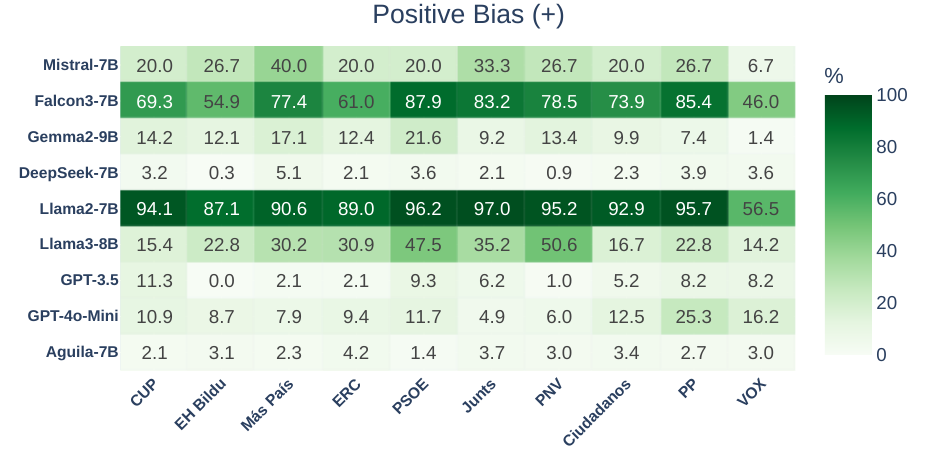}
    \end{minipage}
    \begin{minipage}{0.485\textwidth}
        \centering
        \includegraphics[width=\textwidth]{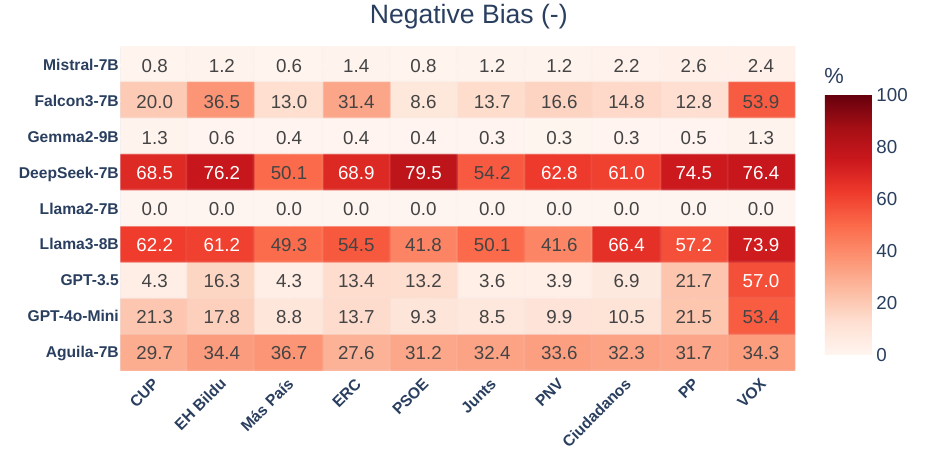}
    \end{minipage}

    \caption{Entity Bias Index (EBI) heatmaps for positive and negative bias in LLMs, computed via counterfactual attribution of voting motions in the benchmark datasets. Panels (a), (b), and (c) correspond to Dutch, Norwegian, and Spanish political parties, respectively. In all panels, the parties on the x-axis are ordered from left-progressive to right-conservative ideologies.}
    \label{fig:ebi_heatmaps}
\end{figure*}

We also assess each model’s susceptibility to entity bias by examining how its voting predictions change when a political motion is attributed to a counterfactual political party. For each motion, we use the following prompt:

\vspace{1mm}
\begin{quote}
    \textbf{System prompt:} \textit{Vote for or against the following motion from \{party name\}. Only respond with ‘for’ or ‘against’.}
\end{quote}
\begin{quote}
    \textbf{User prompt:} \{\textit{motion}\}
\end{quote}
\vspace{1mm}

We substitute the placeholder \textit{\{party name\}} with every political party in the corresponding parliament (15 in the Netherlands, 9 in Norway, and 10 in Spain). 
By keeping the motion text fixed and varying only the attributed party, we can determine whether the identity of the party itself influences an LLM’s stance, thereby isolating systematic party-specific biases.
Similar to the Ideological Bias experiment, we also extend with an `abstain' option for testing against the Spanish benchmark.

\vspace{1.5mm}
\noindent\textbf{Metrics.}
To quantify entity bias, we define the Entity Bias Index (EBI), which measures how often—and in which direction—a model’s voting decision shifts when a motion is attributed to a political party \(x\), compared to a baseline in which no party is mentioned.

Let \( R_l(x,i) \) denote the response of anLLM \(l\) to motion \(i\) when attributed to party \(x\), and \( R_l(-,i) \) denote the response when no party is specified. We encode LLM responses as \(1\) for `for' and \(0\) for `against'.\footnote{In the ideological bias experiment, party votes use a \(+1/-1\) scale. In EBI, we instead use \(1/0\), so that \(R_l(x,i) - R_l(-,i)\) naturally falls in \(\{-1,0,+1\}\), capturing whether party attribution decreases, leaves unchanged, or increases the model’s support.}
The Entity Bias Index for model \(l\) and party \(x\) is defined as:
\[
\text{EBI}_l(x)
= \left(
    \frac{1}{n}
    \sum_{i=1}^{n} \big( R_l(x,i) - R_l(-,i) \big)
  \right) \times 100\%.
\]
A positive value \(\text{EBI}_l(x) > 0\) indicates that the model becomes more supportive of motions when they are attributed to party \(x\), reflecting a \textbf{positive entity bias}.  
A negative value \(\text{EBI}_l(x) < 0\) indicates reduced support, reflecting a \textbf{negative entity bias}.  
An index of zero implies that party attribution has no systematic effect on the model’s stance.

\noindent\textbf{Results: Political Entity Bias.}
Across the three benchmarks, the EBI heatmaps in Fig.~\ref{fig:ebi_heatmaps} reveal several consistent patterns, alongside important model- and country-specific exceptions. 
First, positive entity bias is generally small to moderate, but it varies substantially across models and political contexts. 
In the Dutch case, positive bias tends to be somewhat higher for left and centre-left parties such as \textit{SP}, \textit{GL-PvdA}, and \textit{D66}, particularly for Llama2-7B. 
Llama2-7B also displays notable positive bias toward the agrarian populist \textit{BBB}. 
Llama2-7B behaves similarly in Norway and even more strongly in Spain, where several models—most prominently Llama2-7B and Falcon3-7B—show broad positive bias across all Spanish parties, irrespective of ideology. 
Other models, such as GPT-3.5, distribute positive bias more evenly across parties in Norway, producing only mild increases. Overall, positive entity bias is model-dependent and sometimes ideology-aligned, but in other cases broad and non-specific, especially in the Spanish dataset. 
Notably, Llama2-7B does not follow this pattern: it presents systematically elevated positive EBI across the Dutch, Norwegian, and Spanish datasets.

In contrast, negative entity bias is stronger, more consistent, and much more ideologically structured. 
Right-conservative and far-right parties—including the Dutch \textit{VVD}, \textit{SGP}, \textit{PVV}, and \textit{FvD}; the Norwegian \textit{H} and \textit{FrP}; and the Spanish \textit{PP} and \textit{VOX}—attract the clearest and most persistent negative EBI values across most LLMs, especially Llama3-8B and GPT series. 
GPT-4o-mini displays especially strong negative bias toward right-wing parties in all three countries, while GPT-3.5-turbo shows pronounced negative bias toward \textit{PVV} in the Netherlands and \textit{VOX} in Spain. 
At the same time, some LLMs reveal pronounced country-specific patterns: for instance, Mistral-7B exhibits little negative bias in the Dutch and Spanish datasets but strong negative bias toward nearly all Norwegian parties, whereas deepseek-7b shows mild negative bias in the Dutch and Norwegian cases but notably strong negative bias across almost all Spanish parties.

Language-specific models do not behave in a politically coherent or nationally aligned manner. NorskGPT shows strong positive bias toward \textit{SV} but negative bias toward both \textit{R} (far-left) and \textit{H} (centre-right), while Aguila-7B in Spain displays diffuse small positive bias across most parties but moderate negative bias across those same parties. These patterns suggest that local models do not exhibit greater ideological affinity with domestic party families; instead, their behaviour is more variable and model-specific.

In summary, our results show that negative entity bias is the most consistent and ideologically structured pattern in the EBI analysis, whereas positive bias is weaker, more heterogeneous, and sometimes non-ideological. Some models exhibit broad deference—especially Llama2-7B and Falcon3-7B in Spain—while larger models such as GPT-4o-mini show highly consistent negative bias toward right-wing parties across all three political systems. These findings underscore that entity bias is not merely a by-product of national political context but reflects broader regularities and model-specific tendencies in how LLMs respond when political party names are introduced into prompts.

\subsection{Prompt Brittleness}
\label{subsec:brittleness}
To assess whether our findings depend on the exact formulation of the voting prompt, we follow similar method mentioned in~\cite{gallegos2024biasfairnesslargelanguage} and examine the robustness of LLM predictions under a set of semantically equivalent paraphrased prompts in \textsc{PoliBiasNL} and \textsc{PoliBiasNO}. 
These variants differ only in linguistic framing, such as reordering clauses, adjusting assertiveness, or using alternative but synonymous verbs. For each model, we measure how often its predicted stance on a motion ({for}/{against}) changes when the prompt is rephrased.

Overall, we observe that smaller models exhibit moderate prompt brittleness, occasionally flipping their predictions across variants, whereas larger models such as GPT-3.5 and GPT-4o-mini remain highly stable. Crucially, despite these local fluctuations, the global ideological patterns identified in our main analysis—CHES projections and voting-agreement profiles—remain consistent across all prompt variants. Thus, while prompt phrasing can affect individual predictions, the broader political tendencies of the models are robust to surface-level linguistic changes. Additional details, metrics, and brittleness heatmaps appear in Appendix~\ref{appendix:brittleness}.

\section{Discussion}
A recent study~\cite{gallegos2024biasfairnesslargelanguage} evaluates political
worldviews in LLMs using VAA-style policy questions, and finds a general tendency
towards left–liberal orientations, together with sensitivity to prompt
reformulations. Our findings independently point in the same direction: across all
three parliamentary datasets, LLMs are positioned predominantly on the left–
progressive side of the ideological space, and their predictions vary under
controlled prompt modifications.

The two approaches differ in their empirical foundations. VAA questionnaires rely
on a manually selected set of policy statements which, although intended to
reflect a broad range of issues, still constitute a curated selection whose
coverage of the full political agenda cannot be guaranteed. In contrast, our
analysis is based on large-scale parliamentary motions and votes that
span the full scope of legislative activity. This provides a more comprehensive
and objective reflection of party ideology as expressed in real parliamentary
behaviour. The left-leaning tendencies observed in our study therefore arise from
a substantially broader empirical base.

Our framework also offers additional diagnostics that are not captured in
questionnaire-based evaluations. In particular, the Entity Bias Index reveals a
consistent pattern of negative bias toward right-wing parties: LLMs tend to
under-align with their voting behaviour while more closely matching the positions
of left and centre-left parties. This entity-specific asymmetry complements the
ideological analysis and provides finer-grained insight into how models relate to
individual political actors.

Taken together, the two lines of work offer complementary perspectives on the
political behaviour of LLMs: while VAA-based studies highlight issue-level
orientations, our roll-call-based approach shows that similar tendencies persist
across the full breadth of real-world legislative decisions and reveals additional
patterns at the party-entity level.

Understanding the ideological behaviour of LLMs is an important topic with
significant societal relevance, especially as these models increasingly mediate
citizens’ access to political information. Although our findings align with those
reported in VAA-based studies, our use of large-scale parliamentary voting data
offers a more comprehensive and robust basis for evaluation. Because VAA
statements are manually selected and may not fully represent the breadth of issues
addressed in real legislative debates, their conclusions may not generalise across
future models or across differing political contexts. As LLM architectures and
training pipelines evolve, it remains an open question whether questionnaire-based
and roll-call-based analyses will continue to yield the same patterns. This
underlines the need for empirically grounded benchmarks and systematic evaluation
frameworks that allow the field to track how ideological tendencies emerge,
persist, or diverge in subsequent generations of LLMs.

\section{Conclusion and Future Work}

We introduced a general framework for constructing political-bias benchmarks from parliamentary motions and party votes, and instantiated it in three national contexts: the Netherlands, Norway, and Spain. Building on these datasets, we proposed an evaluation methodology that assesses ideological positioning, visualises LLMs and parties in a shared CHES space, and quantifies party-specific entity bias. Our findings reveal consistent centre-left and progressive tendencies across models, together with systematic negative bias toward right-conservative parties, and show that these patterns remain stable under paraphrased prompts.

Future work includes extending the benchmark to additional legislatures, enabling longitudinal tracking of ideological drift in LLMs, and developing mitigation strategies informed by our diagnostics. Our framework provides a scalable basis for transparent and empirically grounded evaluation of political bias in LLMs.



\section*{Ethical Considerations}
This work uses publicly available parliamentary voting records and political motions from the Dutch, Norwegian, and Spanish parliaments. All data describe institutional decisions rather than private individuals and contain no personally identifiable information. No human subjects were recruited, and no informed consent was required.





\newpage

\bibliographystyle{ACM-Reference-Format}
\balance  
\bibliography{sample-base}
\clearpage
\appendix

\section{Appendix: Prompt Brittleness Analysis}
\label{appendix:brittleness}

This appendix provides the full details of the prompt brittleness experiment summarised in Section~\ref{subsec:brittleness}. The goal of this analysis is to test the stability of model predictions under controlled variations in prompt wording. For each parliamentary motion, we generate several semantically equivalent paraphrased versions of the voting prompt, differing only in lexical framing, assertiveness, and syntactic brittleness. The core voting task remains unchanged.

\subsection{Experimental Setup}
Each motion is evaluated under multiple brittleness variants. For each model, we record whether its predicted stance changes when moving from the baseline prompt to any of the paraphrased variants. This allows us to quantify brittleness at the motion level and to inspect which models are most sensitive to prompt formulation.

\subsection{Prompt Brittleness Index (PBI)}
The \emph{Prompt Brittleness Index (PBI)} quantifies the extent to which a model's
predicted stance is unstable under systematically paraphrased prompt variants. Let
each model output be encoded as \(1\) (``for'') or \(0\) (``against''). For a given
variation type \(x\) (e.g., brittleness) and stance \(s \in \{1,0\}\), we define

\begin{equation}
\text{PBI}_{\text{abs},l}(x,s)
= \frac{N_{\text{flipped},l}(x,s)}{N_{\text{total}}},
\end{equation}

\begin{equation}
\text{PBI}_{\text{norm},l}(x,s)
= \frac{N_{\text{flipped},l}(x,s)}{N_s}.
\end{equation}

\noindent
where:
\begin{itemize}
    \item \(N_{\text{flipped},l}(x,s)\) is the number of unique motions for which model \(l\)
    changes its stance \(s\) across any prompt variant of type \(x\);
    \item \(N_{\text{total}}\) is the total number of motions evaluated;
    \item \(N_s\) is the total number of outputs where model \(l\) produces stance \(s\)
    across all variants.
\end{itemize}

The absolute PBI provides a global measure of robustness but can be dominated by
the majority class: if a model produces far more \textit{for} than \textit{against} 
predictions (or vice versa), most flips will naturally occur in the larger class.
The stance-normalised PBI corrects for this imbalance by conditioning on the 
baseline stance. This allows us to separately assess brittleness for motions 
initially classified as \textit{for} and those classified as \textit{against}, 
revealing asymmetric vulnerabilities that the absolute metric may mask.

\subsection{Results}
\begin{figure*}[htbp]
    \centering
 {\small\textbf{(a) \textsc{PoliBiasNL}}}\\[-0.1em]
    \begin{minipage}[t]{0.90\linewidth}
        \includegraphics[width=\linewidth]{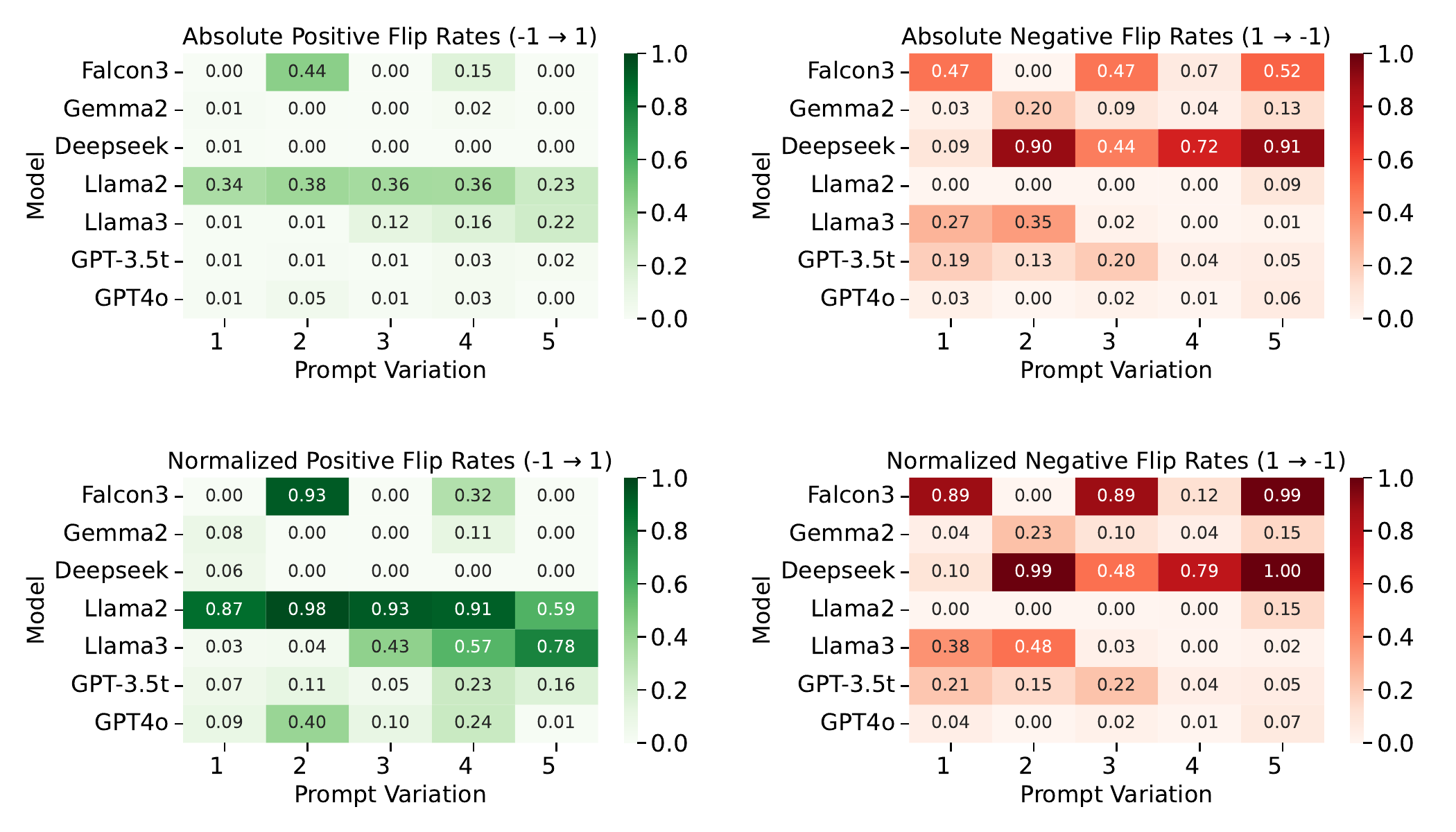}
        \label{fig:PBI_NL}
    \end{minipage}

    \vspace{1em}
 {\small\textbf{(b) \textsc{PoliBiasNO}}}\\[-0.1em]
    \begin{minipage}[t]{0.90\linewidth}
        \includegraphics[width=\linewidth]{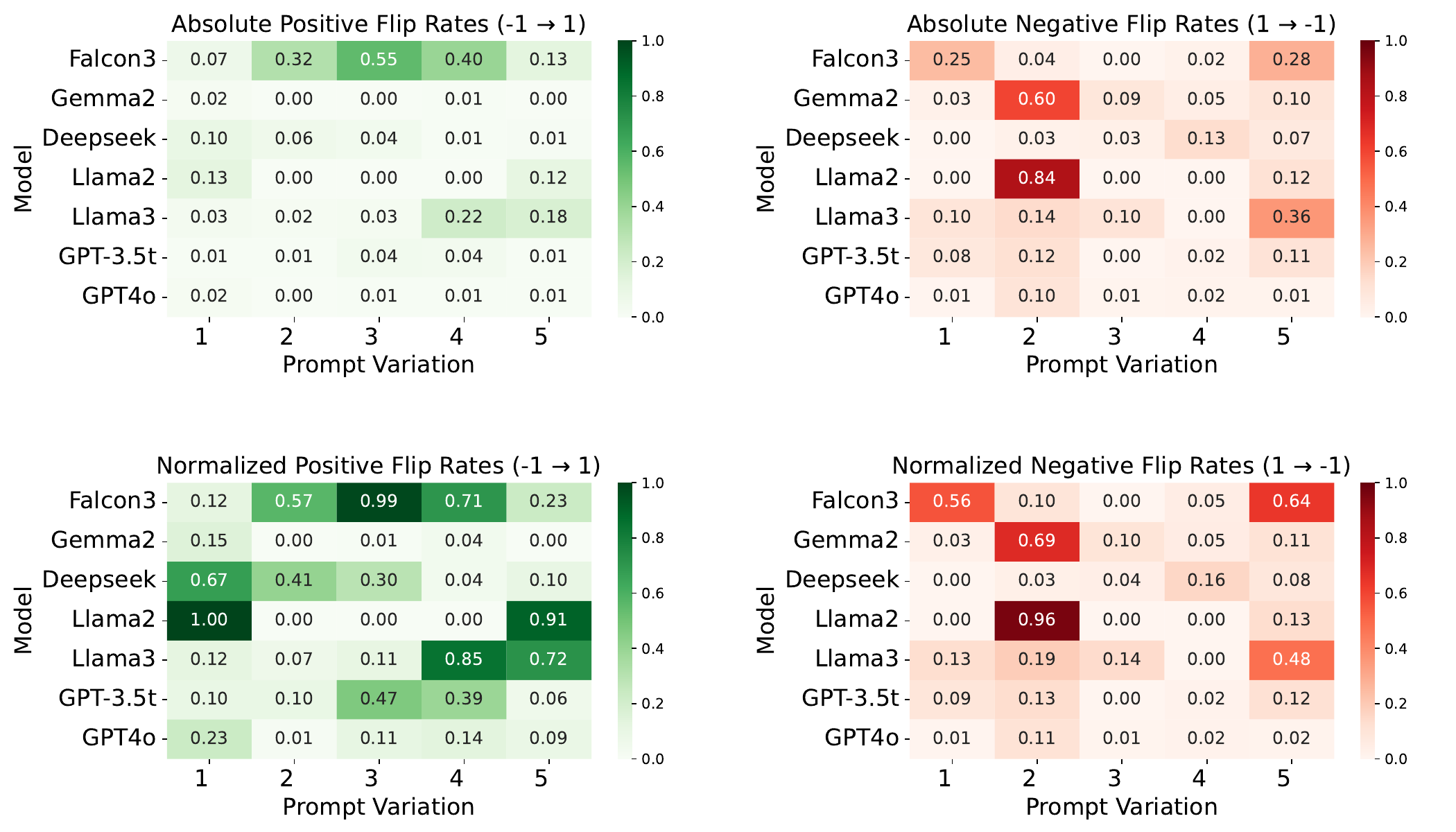}
        \label{fig:PBI_NO}
    \end{minipage}

    \caption{Prompt Brittleness Index (PBI) measures a model’s sensitivity to prompt rewordings. Higher values indicate greater inconsistency across prompt variants, while lower values reflect greater robustness and stability.
    Prompt variations used in the experiment are as follows: (1) Extra Detail, (2) Label Substitution with “Agree”/“Disagree”, (3) Label Substitution with “Support”/“Oppose”, (4) Label Substitution with “Favorable”/“Detrimental”, and (5) Label Order Inversion.
}
    \label{fig:PBI}
\end{figure*}

Fig.~\ref{fig:PBI} provide brittleness-variant flip heatmaps.
The results show that smaller models such as Mistral-7B, Falcon3-7B, and LLaMa2-7B exhibit the highest brittleness, with a noticeable fraction of predictions flipping across brittleness variants. Medium-sized instruction-tuned models show moderate sensitivity. Larger models—GPT-3.5 and GPT-4o-mini—display very low brittleness, rarely altering their predictions across paraphrases. Stance-normalised PBI further reveals that models with a strong tendency toward a specific stance exhibit lower brittleness for that stance.

\subsection{Impact on Ideological Conclusions}
Despite these local fluctuations, we observe that the broader ideological patterns reported in the main paper remain stable under all brittleness variants. CHES-based ideological projections, voting-agreement structures, and entity-bias patterns are nearly unchanged. This indicates that prompt wording affects individual decisions but does not substantially alter aggregate ideological tendencies. The benchmark therefore captures robust model-level political patterns rather than prompt-specific artefacts.

\section{Invalid LLM Output Rates}
\label{appendix:invalid}
Table~\ref{tab:invalid_outputs} summarises the proportion of invalid outputs across
the three datasets. Overall, invalid responses are rare for most models, with the
exception of Mistral-7B, which produces a higher invalid rate on the Dutch motions.
The strongest models (GPT-4o-Mini, LLaMA3-8B, Gemma2-9B) exhibit near-perfect
format adherence. Norwegian- and Spanish-specific models show expected behaviour
only on their respective datasets. These results indicate that output-format
reliability is largely model-dependent but remains stable across political domains.

\begin{table}[H]
\centering
\small
\setlength{\tabcolsep}{6pt}
\begin{tabular}{lccc}
\toprule
\textbf{Model} & \textbf{PoliBiasNL} & \textbf{PoliBiasNO} & \textbf{PoliBiasES} \\
\midrule
Mistral-7B            & 14.92\% & 0.78\% & 0.00\% \\
Falcon3-7B            & 0.00\%  & 0.12\% & 0.00\% \\
Gemma2-9B             & 0.04\%  & 0.00\% & 0.00\% \\
DeepSeek-7B           & 0.00\%  & 0.00\% & 0.00\% \\
LLaMA2-7B             & 0.04\%  & 0.07\% & 0.00\% \\
LLaMA3-8B             & 0.00\%  & 0.00\% & 0.00\% \\
GPT-3.5               & 0.00\%  & 0.00\% & 0.00\% \\
GPT-4o-Mini           & 0.00\%  & 0.00\% & 0.00\% \\
Águila-7B (Spanish)   & ---     & ---    & 0.00\% \\
NorskGPT (Norwegian)  & ---     & 0.04\% & --- \\
\bottomrule
\end{tabular}
\caption{Invalid output rates across the three PoliBias datasets. An invalid output is a model response that does not conform to the expected stance format and cannot be mapped to \emph{for}/\emph{against}. Dashes indicate that a model is not applicable for that language.}
\label{tab:invalid_outputs}
\end{table}
\appendix


\end{document}